\documentclass[lettersize,journal]{IEEEtran}
\usepackage{amsmath,amsfonts,amssymb}
\usepackage{array}
\usepackage[caption=false,font=footnotesize,labelfont=rm,textfont=rm]{subfig}
\usepackage{textcomp}
\usepackage{stfloats}
\usepackage{url}
\usepackage{verbatim}
\usepackage{graphicx}
\usepackage{cite}
\usepackage{color}

\usepackage{tikz}
\usepackage{comment}

\usepackage[ruled]{algorithm2e}
\usepackage{multirow}
\newcommand{\tabincell}[2]{\begin{tabular}{@{}#1@{}}#2\end{tabular}} 
\usepackage{booktabs}
\usepackage{bm}
\newcommand \ie {$i$.$e$.}
\newcommand \eg {$e$.$g$.}

\begin{document}

\title{PSAQ-ViT V2: Towards Accurate and General Data-Free Quantization for Vision Transformers} 

\author{Zhikai Li, Mengjuan Chen, Junrui Xiao, and Qingyi Gu, \IEEEmembership{Member, IEEE}
	\thanks{This work was supported {\color{black} in part by the National Key Research and Development Program of China under Grant  2022ZD0119402; in part by the National Natural Science Foundation of China under Grant 62276255.} 
	 \emph{(Corresponding author: Qingyi Gu.)} }
	\thanks{
		Z. Li, M. Chen, J. Xiao and Q. Gu are with the Institute of Automation, Chinese Academy of Sciences, Beijing 100190, China, and Z. Li and J. Xiao are also with the School of Artificial Intelligence, University of Chinese Academy of Sciences, Beijing 100049, China (e-mail: lizhikai2020@ia.ac.cn; chenmengjuan2016@ia.ac.cn; xiaojunrui2020@ia.ac.cn; qingyi.gu@ia.ac.cn).
	}
}



\maketitle

\begin{abstract}
	Data-free quantization can potentially address data privacy and security concerns in model compression, and thus has been widely investigated. Recently, PSAQ-ViT {\color{black}(paper entitled ``Patch Similarity Aware Data-Free Quantization for Vision Transformers")} designs a relative value metric, patch similarity, to generate data from pre-trained vision transformers (ViTs), achieving the first attempt at data-free quantization for ViTs. In this paper, we propose PSAQ-ViT V2, a more accurate and general data-free quantization framework for ViTs, built on top of PSAQ-ViT. More specifically, following the patch similarity metric in PSAQ-ViT, we introduce an adaptive teacher-student strategy, which facilitates the constant cyclic evolution of the generated samples and the quantized model (student) in a competitive and interactive fashion under the supervision of the full-precision model (teacher), thus significantly improving the accuracy of the quantized model. Moreover, without the auxiliary category guidance, we employ the task- and model-independent prior information, making the general-purpose scheme compatible with a broad range of vision tasks and models. 
	Extensive experiments are conducted on various models on image classification, object detection, and semantic segmentation tasks, and PSAQ-ViT V2, with the naive quantization strategy and without access to real-world data, consistently achieves competitive results, showing potential as a powerful baseline on data-free quantization for ViTs. For instance, with Swin-S as the (backbone) model, 8-bit quantization reaches 82.13 top-1 accuracy on ImageNet, 50.9 box AP and 44.1 mask AP on COCO, and 47.2 mIoU on ADE20K.
	We hope that accurate and general PSAQ-ViT V2 can serve as a potential and practice solution in real-world applications involving sensitive data.
	Code is released and merged at: \url{https://github.com/zkkli/PSAQ-ViT}.

\end{abstract}


\section{Introduction}
\IEEEPARstart{T}{hanks} to the great success in a variety of vision applications, vision transformers (ViTs) have recently drawn widespread attention in both research and practice \cite{khan2021transformers,han2020survey,kaselimi2022vision,alfasly2022effective}. Nevertheless, their large memory footprints, computational overheads, and power consumption are challenging for real-world applications \cite{tang2021patch,jia2021efficient,lin2021fq}, especially for the deployment and efficient inference on resource-constrained edge devices \cite{liu2021post,Li2022IViTIQ}. Model quantization \cite{krishnamoorthi2018quantizing,gholami2021survey}, which reduces the representation precision of parameters, is an effective procedure to reduce model complexity in a hardware-friendly manner \cite{zhang2018lq,elthakeb2020gradient,chin2020one,hubara2016binarized,rastegari2016xnor,kim2020exploiting,fei2021general,wang2020unsupervised}. Unfortunately, existing quantization methods require access to the original training datasets to calibrate/fine-tune the quantization parameters \cite{li2022patch}, which can raise widely-held data privacy and security issues and thus cannot be applied in data-sensitive scenarios \cite{cai2020zeroq,zhang2021diversifying,guo2022squant,yin2020dreaming}.

Consequently, data-free quantization, which works to generate fake data from the prior information of the pre-trained model itself to replace the real data used in the quantization process, has been widely investigated and explored \cite{cai2020zeroq,xu2020generative,zhang2021diversifying,zhong2021intraq,choi2021qimera,guo2022squant}. Existing approaches have focused on convolutional neural networks (CNNs), which utilize batch normalization (BN) regularization to facilitate the distribution of the generated samples to match the real-data statistics embedded in the BN layers of the pre-trained full-precision (FP) model \cite{cai2020zeroq,xu2020generative,zhang2021diversifying}. However, this is not applicable to ViTs with layer normalization (LN), because LN is dynamically computed and does not store the statistical prior of the original data \cite{li2022patch}. Thus, designing specific data-free quantization schemes for ViTs is highly desired.

To this end, PSAQ-ViT {\color{black}(paper entitled ``Patch Similarity Aware Data-Free Quantization for Vision Transformers")} \cite{li2022patch} makes a first attempt to quantize ViTs without any real-world data, filling the gap in the community.
Since there is no available \emph{absolute value} metric in ViTs like BN statistics, PSAQ-ViT discovers a general difference in the self-attention module's processing of Gaussian noise and real images, \ie, patch similarity, and accordingly develops a \emph{relative value} metric to reduce this difference and thus optimize Gaussian noise to approximate real images.
Despite achieving good results, the accuracy and generality of this scheme are still less than expected.
First, sample generation and model quantization are considered as two independent phases, which hinders the informativeness and diversity of the generated samples and the on-going learning of the quantized model.
Second, it utilizes the auxiliary category prior to enhance the semantic features of the class-conditional samples, which unfortunately restricts the scheme to the image classification task and makes it inapplicable to high-level vision applications such as object detection and semantic segmentation.

\begin{figure*}
	\centering
	\includegraphics[width=0.75\linewidth]{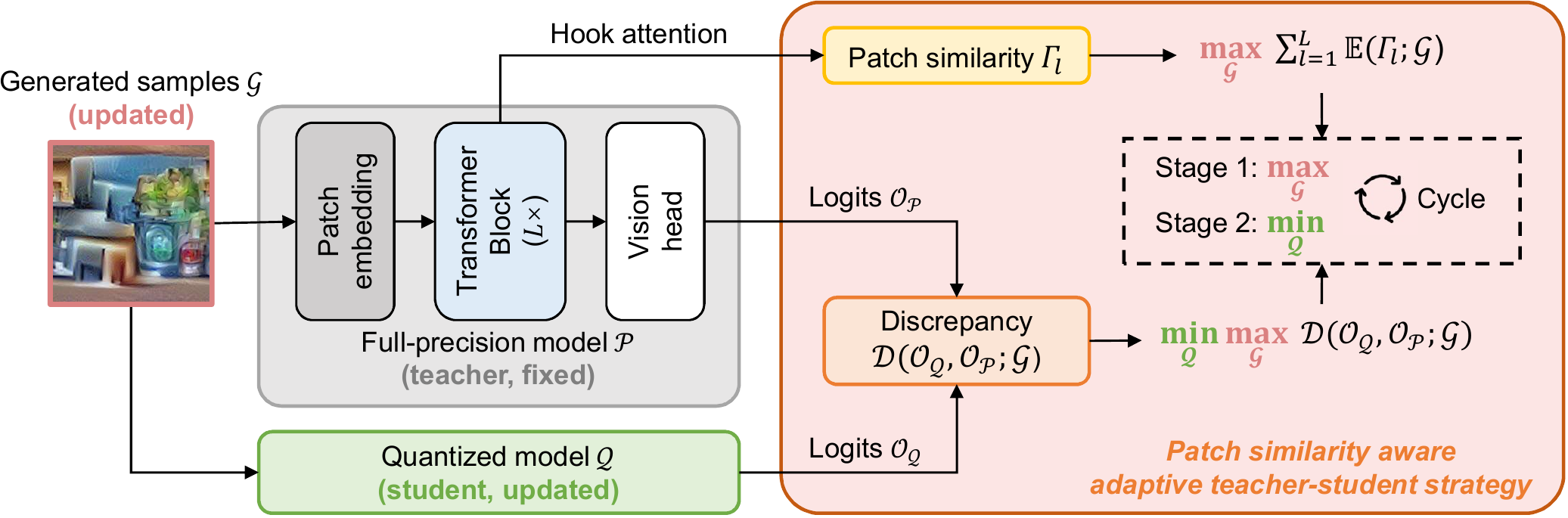}
	\caption{{\color{black}Overview of the proposed PSAQ-ViT V2 framework.} Building on patch similarity proposed in PSAQ-ViT, we introduce an adaptive teacher-student strategy to enhance the performance, where the generated samples and the quantized model cyclically play a minimax game with respect to the model discrepancy. Specifically, the FP model is kept fixed; in stage 1 of each cycle, the diversity of patch similarity and the model discrepancy are maximized to update the generated samples, and in stage 2, the model discrepancy is minimized to update the quantized model. The alternation of these two stages enables the constant cyclic evolution of the generated samples and the quantized model.}
	\label{fig:overview}
\end{figure*}

To address the above issues, we propose PSAQ-ViT V2, an enhanced version to enable more \emph{accurate} and \emph{general} data-free quantization for ViTs. This is achieved by well-designed refinements building on patch similarity in PSAQ-ViT, as shown in Fig. \ref{fig:overview}. 
Specifically, we introduce a teacher-student strategy that facilitates the evolution of the quantized model using the generated samples under the supervision of the FP model. It is worth noting that instead of a one-time synthesis of all samples, the above procedure is adaptive, \ie, the generated samples and the quantized model cyclically play a minimax game with respect to the model discrepancy. This competitive and interactive style forces the progressive emergence of new and diverse features in the generated samples, thus promoting constant and effective learning of the quantized model.
In this process, image augmentation is used to cost-effectively expand the samples in each mini-batch.
Moreover, thanks to the removal of the category prior, the objective functions in this work are independent to the tasks and models, whose general-purpose nature makes the scheme compatible with a broad range of vision tasks, including image classification, object detection, and semantic segmentation, and also allows for different teacher and student model structures, \eg, the learning of quantized DeiT-T can be performed under the supervision of DeiT-B.

To sum up, our contributions are as follows:

\begin{itemize}
	\item On top of the relative value metric (patch similarity) for sample generation in PSAQ-ViT, we propose an enhanced version, which is a more accurate and general data-free quantization framework for ViTs.
	\item We introduce an adaptive teacher-student strategy, whose competitive and interactive properties allow to progressively force new sample features to emerge, thus facilitating the constantly-evolving quantized model under the supervision of the FP model.
	\item The prior information used in this scheme is task- and model-independent, just in contrast to the category prior, making the scheme general-purpose and compatible with various vision tasks and models.
	\item Extensive experiments are performed with various models on benchmark datasets for multiple vision tasks, and the results demonstrate notable advantages in accuracy and generality, with the potential to serve as a strong baseline.
\end{itemize}

\section{Related Works}
\subsection{Vision Transformers}
With the success in natural language processing applications, transformers, which enjoy the global receptive field offered by the self-attention mechanism, have recently achieved significant performance gains on a range of computer vision applications \cite{arnab2021vivit,zhou2021deepvit,chen2021transformer,wu2021rethinking,han2021transformer}.
By reshaping the image into a sequence of flattened 2D patches as the input, ViT \cite{dosovitskiy2020image} is the first work to apply the transformer-based models to the computer vision community, which boosts the performance baseline on the image classification task.
DeiT \cite{touvron2021training} employs a distillation token to allow the student to learn from the teacher through attention, which can achieve competitive results on ImageNet with low training data cost. 
Swin \cite{liu2021swin} constructs hierarchical feature maps using the shift of the window partition between consecutive self-attention layers and has linear computational complexity to image size, further demonstrating the potential of transformer-based models as vision backbone.
In addition to the image classification task, ViTs are also emerging as popular solutions for other high-level vision tasks, such as object detection \cite{carion2020end,zhu2020deformable}, semantic segmentation \cite{chen2021pre}, and video recognition \cite{neimark2021video}. 

Unfortunately, despite the excellent performance, ViTs typically have complex model architectures and their high number of parameters and computation is intolerable for real-world applications \cite{tang2021patch,liu2021post}. To this end, several works recently focus on lightweight ViTs, such as MobileViT \cite{mehta2021mobilevit} and TinyViT \cite{wu2022tinyvit}, whose model complexity, however, is still far from satisfactory, especially for the deployment on the resource-constrained edge devices.
As a result, model compression techniques for ViTs are regarded as necessary and promising solutions.

\subsection{Data-Driven Quantization}
Model quantization, which replaces 32-bit floating-point parameters with low-bit values, is a popular and promising way to reduce the complexity of neural networks \cite{krishnamoorthi2018quantizing,gholami2021survey}, thus allowing their deployment and real-time inference on edge devices. To date, there have been considerable researches attempting to quantize CNNs. A selection of notable works performs quantization-aware training (QAT) to improve the accuracy of the quantized model \cite{zhang2018lq,li2019additive,choi2018pact,esser2019learned}, where the straight-through estimator (STE) \cite{bengio2013estimating} is used to approximate the gradient back-propagation of discrete parameters. To reduce the computational cost of QAT, other efforts propose post-training quantization (PTQ) methods \cite{jacob2018quantization,li2019fully,choukroun2019low,wu2020easyquant,nagel2020up,li2021brecq}, which focus on the calibration of quantization parameters. 

In addition, several quantization schemes designed for ViTs have been recently presented. 
Ranking-Aware \cite{liu2021post} introduces a ranking loss into the conventional quantization objective to keep the relative order of the self-attention results after quantization.
PTQ4ViT \cite{yuan2021ptq4vit} proposes the twin uniform quantization method to reduce the quantization error on the activation values after Softmax and GELU functions.
FQ-ViT \cite{lin2021fq} designs specific quantization strategies for Softmax and LN components to obtain the fully quantized ViTs.
I-ViT \cite{Li2022IViTIQ} utilizes integer bit-shifting to approximate Softmax and GELU operations to achieve efficient integer-only inference for ViTs.

However, all the above methods rely heavily on the original training data, making them inapplicable in data-sensitive scenarios where the original data is not available.

\subsection{Data-Free Quantization}
With the ability to compress models without access to any real data, data-free quantization can potentially address data privacy and security issues and is thus receiving increasing attention \cite{guo2022squant,choi2021qimera}.
ZeroQ \cite{cai2020zeroq} proposes BN regularization to facilitate the distribution matching between the generated samples and the real data, and then uses the generated samples to calibrate the quantization parameters.
Based on BN regularization, DSG \cite{zhang2021diversifying} utilizes slack distribution alignment and layerwise sample enhancement to improve the diversity of the generated samples.
GDFQ \cite{xu2020generative} and IntraQ \cite{zhong2021intraq} combine the category prior with BN statistics to generate class-conditional samples while matching the real-data distribution, powering the performance of data-free quantization. 
These methods, however, are only applicable to CNNs and not to ViTs, because LN in ViTs, unlike BN, does not store any prior information of the original training data.

To address the above challenge, {\color{black}PSAQ-ViT \cite{li2022patch}, as the previous version of this work,} carries out an insight into the general difference in the self-attention module's processing of Gaussian noise and real images, and thus proposes a \emph{relative value} metric, patch similarity, to optimize the Gaussian noise to approximate real images.
In this work, building on patch similarity, we are interested in making the scheme more accurate and general by introducing advanced quantization learning strategies, which further advances the data-free quantization community for ViTs.

\section{Methodology}

\textbf{Framework Overview:}
Fig. \ref{fig:overview} illustrates an overview of the proposed framework. The whole scheme requires only a pre-trained FP model, without any other information especially real-world data, to obtain a quantized model with superior performance.
In this scheme, following PSAQ-ViT, patch similarity remains an essential component in driving Gaussian noise to approximate the real images, as detailed in Section \ref{sec:patch_similarity}.
On this basis, we introduce an adaptive teacher-student strategy, which facilitates consistent and effective learning of the quantized model using the generated samples with constantly new features under the supervision of the FP model. This is achieved by the competition and interaction between the generated samples and the quantized model, and specifically, they cyclically play a two-player minimax game with respect to the model discrepancy, as detailed in Section \ref{sec:adaptive}.
The overall pipeline described above is summarized and presented in Section \ref{sec:pipeline}.

\subsection{Preliminaries}
Standard ViTs consist of an embedding layer and several stacked transformer blocks. First, the embedding layer reshapes the input image into a sequence of flatted 2D patches, which are then linearly projected into a $d$-dimensional space to obtain the token embeddings $X\in \mathbb{R}^{N\times d}$. Here, $N$ is the number of patches.

The token embeddings $X$ are then fed into a series of transformer blocks, where each block is composed of a multi-head self-attention (MSA) module and a multi-layer perceptron (MLP) module as follows:
\begin{equation}
	\begin{aligned}
		\hat{X} = X+\text{MSA}(\text{LN}(X)) \\
		Y = \hat{X}+\text{MLP}(\text{LN}(\hat{X}))
	\end{aligned}
\end{equation}

MSA calculates the attention between different patches to learn inter-patch representations from a global view as follows:
\begin{equation}
	\begin{aligned}
		\text{MSA}(X)&=\text{Concat}(\text{head}_1,\cdots,\text{head}_H)W^o \\
		\text{where head}_i = & \; \text{Attn}(Q_i,K_i,V_i)=\text{softmax}(\frac{Q_iK_i^T}{\sqrt{d}})V_i
	\end{aligned}
\end{equation}
where $H$ is the number of attention heads. Here, query $Q_i$, key $K_i$, and value $V_i$ are computed by linear projections using matrix multiplication, $i$.$e$., $Q_i=XW_i^Q$, $K_i=XW_i^K$, $V_i=XW_i^V$.
Then, MLP employs two dense layers to process the obtained attention for high-dimensional feature mapping and information fusion as follows:
\begin{equation}
	\text{MLP}(\hat{X})=\text{GELU}(\hat{X}W_1+b_1)W_2+b_2
\end{equation}

Finally, the blocks are followed by the classification/detection/segmentation heads depending on the vision task to get the final result.

In this paper, we work on quantizing all the parameters (weights and activations) of the operations (\eg, large matrix multiplication) in the embedding layer, transformer blocks, and subsequent vision heads, including the input and output. For the quantization strategy, we employ the simple uniform quantization, which is the most popular and hardware-friendly method and is defined as follows:
\begin{equation}
	\theta^q = \lfloor\frac{\text{clip}(\theta^p,q_0,q_{2^k-1})-q_0}{\Delta}\rceil,\,\text{where}\,\Delta = \frac{q_{2^k-1}-q_0}{2^k-1}
	\label{eq:quant}
\end{equation}
where $\theta^p$ and $\theta^q$ denote the parameters of the FP model and the quantized model, respectively. Here, $\lfloor\cdot\rceil$ is the round operator, $q_0$ and $q_{2^k-1}$ are the quantization clipping values, and $k$ is the quantization bit-precision. 

\subsection{{\color{black}PSAQ-ViT: Patch Similarity Metric}}
\label{sec:patch_similarity}
As stated before, ViTs do not have BN layers that store information about the original training data, leaving no available absolute value prior information to generate the samples for performing quantization, which is the major challenge of data-free quantization for ViTs.
Therefore, {\color{black}the previous version of this work, PASQ-ViT, is} to mine deeper into the prior information of the pre-trained ViTs and thus explore a reliable \emph{relative value} metric that can well describe the general difference between Gaussian noise and real images, so that we can reduce this difference to make the Gaussian noise approximate the real images.

\begin{figure*}
	\centering
	\includegraphics[width=0.9\linewidth]{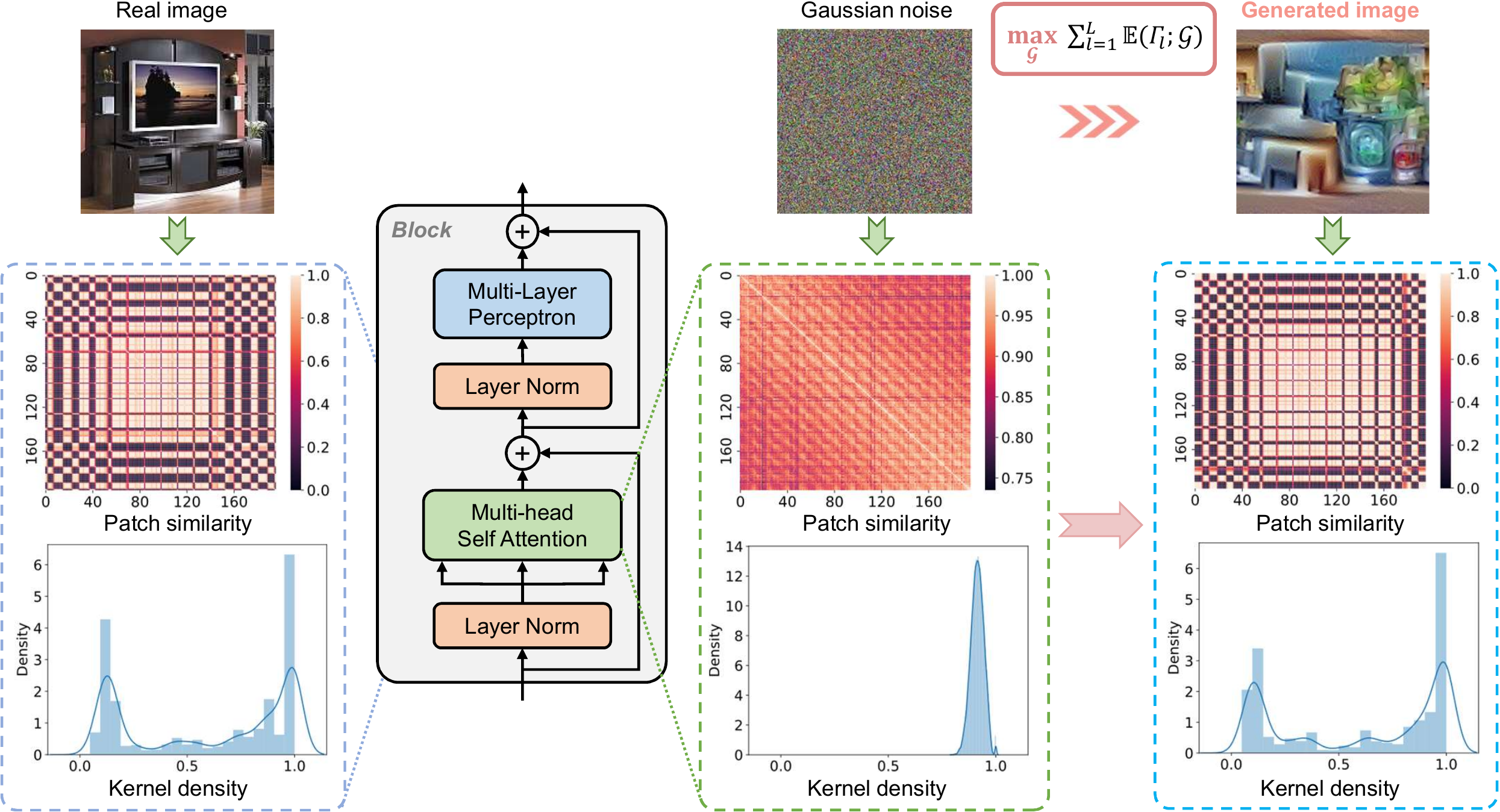}
	\caption{Illustration of the patch similarity-aware sample generation approach via maximizing $\sum_{l=1}^{L}\mathbb{E}(\bm{\Gamma}_l;\mathcal{G})$. When the input is Gaussian noise, patches are grouped into one category (foreground or background), leading to homogeneous patch similarity and a unimodal kernel density curve. Our generated image can potentially represent the real-image features, producing diverse patch similarity and a bimodal kernel density curve, where the left and right peaks describe inter- and intra-category similarity, respectively.}
	\label{fig:patch_similarity}
\end{figure*}

As the unique structure in ViTs, the self-attention module learns inter-patch feature representations, which are believed to contain a certain amount of original data information.
Consequently, we provide an in-depth analysis of the inference process of the self-attention module, and then we observe that the reason the model can make good decisions is that the self-attention module can distinguish the foreground from the background of the training data, thus allocating more attention to the foreground that is more important for the decision. Since the input of ViTs are independent vectors mapped by 2D patches, the responses of the self-attention module to different patches are significantly different, $i$.$e$., the foreground patches receive more attention.
More specifically, when the pre-trained model executes inference, real images consistently produce the above features, while Gaussian noise, whose foreground is not easily extracted, does not have a similar capability and inevitably leads to homogeneous responses, as shown in Fig. \ref{fig:patch_similarity}. Note that the real images here are only used to verify the general difference ($e$.$g$., a certain metric of the real images is always larger than that of Gaussian noise), and they will not be involved in any subsequent process.
Therefore, this general difference can indirectly represent the prior information of ViTs and thus can be used to design the relative value metric to guide the sample generation.

Based on the above insights, we work on designing a reliable metric that can measure
the diversity of the self-attention module's responses. 
For the $l$-th block in ViTs, the output of the MSA module is defined as $O_l \in \mathbb{R}^{B\times H\times N\times d}$ ($l\in \{1,\cdots,L\}$), where each dimension denotes the batch size, number of heads, number of patches, and hidden size, respectively.
To simplify the expression, we ignore the batch dimension, $i$.$e$., $O_l \in \mathbb{R}^{H\times N\times d}$.

Due to the relative value metric, it is necessary to first normalize $O_l$ to ensure the fairness of the comparison. We accomplish this by calculating the cosine similarity between each subspace vector in the patch dimension, specifying the data range at [-1, 1], as follows:
\begin{equation}
	\Gamma_l(u_i,u_j) = \frac{u_i\cdot u_j}{||u_i||\;||u_j||}
	\label{eq:gamma}
\end{equation}
where the numerator represents the inner product of the vectors, and $||\cdot||$ denotes the $l_2$ norm. Here, $u_i,u_j \in \mathbb{R}^{H\times d}$ ($i,j\in \{1,\cdots, N\}$) is the $i$-th/$j$-th vector in the patch dimension of $O_l$, and $\Gamma_l(u_i,u_j)$ represents the cosine similarity between $u_i$ and $u_j$.
After pairwise calculations, we obtain the $l$-th block's cosine similarity matrix $\bm{\Gamma}_l=[\Gamma_l(u_i,u_j)]_{N\times N}$, which is a symmetric matrix and is termed as \emph{patch similarity}. The diversity of patch similarity can potentially represent the diversity of the original data, which not only elegantly achieves data normalization, but also has the additional advantage of achieving reasonable $\frac{Hd}{N}$-fold  dimensionality reduction ($\mathbb{R}^{H\times N\times d}\to \mathbb{R}^{N\times N}$). For instance, for the DeiT-B model, the amount of data is reduced by a factor of 3.92, which can greatly improve the subsequent computational efficiency.

Then, the diversity of patch similarity is measured by the information entropy, which can represent the amount of information expressed by the input data.
{\color{black}To ensure gradient back-propagation, we employ the differential entropy that has a continuous nature.} More intuitively, the $l$-th block's differential entropy of patch similarity $\bm{\Gamma}_l$ at the input generated data $\mathcal{G}$ is calculated as follows:
\begin{equation}
	\mathbb{E}(\bm{\Gamma}_l;\mathcal{G}) = -\int f_h(x) \cdot \log \left[ f_h(x) \right] dx
\end{equation}
where $f_h(x)$ is the continuous probability density function of $\bm{\Gamma}_l$, which is obtained using kernel density estimation as follows:
\begin{equation}
	f_h(x) = \frac{1}{M}\sum_{m=1}^{M}K_h(x-x_m)=\frac{1}{Mh}\sum_{m=1}^{M}K(\frac{x-x_m}{h})
\end{equation}
where $K(\cdot)$ is the kernel ($e$.$g$. normal kernel), $h$ is the bandwidth, $x_m$ ($m\in \{1,\cdots,M\}$) is a training point drawn from $\bm{\Gamma}_l$ and is the center of a kernel, and $x$ is the given test point.

Finally, we sum the differential entropy of each block to account for the diversity of patch similarity across all blocks, and the \emph{Patch Similarity Entropy Loss} is defined as follows:
\begin{equation}
	\mathcal{L}_{PSE} = \sum_{l=1}^{L}\mathbb{E}(\bm{\Gamma}_l;\mathcal{G})
	\label{eq:l_pse}
\end{equation}

\subsection{Adaptive Teacher-Student Strategy}
\label{sec:adaptive}
To enhance the performance of data-free quantization in PSAQ-ViT, we introduce an adaptive teacher-student strategy, which drives the quantized model (student) to imitate the FP model (teacher) with the help of the generated samples. Here, adaptivity is reflected in the fact that instead of a one-time synthesis, the generated samples evolve alternately with the quantized model in a cycle. This competitive and interactive nature forces the gradual emergence of new features in the generated samples, and thus facilitates the quantized model to constantly learn new knowledge. Specifically, they cyclically play a two-player minimax game with respect to the model discrepancy $\mathcal{D}(\mathcal{O}_\mathcal{Q},\mathcal{O}_\mathcal{P};\mathcal{G})$ as follows:
\begin{equation}
	\mathcal{Q}^*,\mathcal{G}^* = \arg \min_\mathcal{Q} \max_\mathcal{G} \mathcal{D}(\mathcal{O}_\mathcal{Q},\mathcal{O}_\mathcal{P};\mathcal{G})
	\label{eq:minimax}
\end{equation}
where $\mathcal{O}_\mathcal{Q}$ and $\mathcal{O}_\mathcal{P}$ are the output logits of the quantized model and the FP model when the input is $\mathcal{G}$, respectively.

In the typical real-data-driven knowledge transfer or distillation applications, Kullback-Leibler (KL) divergence is a popular metric for estimating model discrepancy. However, our minimax game is different from these cases where the student can greedily perform one-time learning from the soft targets produced by the teacher, thus it has unsatisfactory contributions. Specifically, the distance described by KL divergence is loose, which makes the optimization space become small after the quantized model converges to the currently generated samples, \ie, the results of KL divergence are similar when inputting samples with new features, which causes gradient decay and thus hinders the competition and constant evolution of the generated samples and the quantized model.
Thus, we employ a more robust function, Mean Absolute Error (MAE), to construct a better search space, and the \emph{Discrepancy Loss} is defined as follows:
\begin{equation}
	\mathcal{L}_\mathcal{D} = \mathcal{D}(\mathcal{O}_\mathcal{Q},\mathcal{O}_\mathcal{P};\mathcal{G}) = \frac{1}{n}||\mathcal{Q}(\mathcal{G})-\mathcal{P}(\mathcal{G})||_1
	\label{eq:l_d}
\end{equation}
where $n$ is the length of the output logits, \eg, category number for classification and prediction map size for detection and segmentation.


In practice, each cycle in the above minimax game can be divided into two stages, each of which is discussed in detail below. In stage 1, we fix the FP model $\mathcal{P}$ and the quantized model $\mathcal{Q}$ and only update the generated samples $\mathcal{G}$ by maximizing the model discrepancy $\mathcal{D}(\mathcal{O}_\mathcal{Q},\mathcal{O}_\mathcal{P};\mathcal{G})$. Intuitively, the purpose of this action is to encourage generating more confusing samples. Thus, for this purpose, new features are forced to emerge gradually in the generated samples $\mathcal{G}$, which can drive the samples to discover missing representations and potentially increase their informativeness and diversity.
In stage 2, the FP model $\mathcal{P}$ and the generated samples $\mathcal{G}$ are fixed, and the model discrepancy $\mathcal{D}(\mathcal{O}_\mathcal{Q},\mathcal{O}_\mathcal{P};\mathcal{G})$ is minimized to update the quantized model $\mathcal{Q}$, which drives the quantized model to imitate the FP model to obtain the promising performance.
In this manner, the two stages are executed alternately, allowing for the cyclic evolution of the generated samples and the quantized model.





\subsection{The Overall Pipeline}
\label{sec:pipeline}

\begin{algorithm}[t]
	\caption{The PSAQ-ViT V2 Pipeline}
	\label{alg:PSAQ}
	\SetAlgoLined
	\KwIn{A pre-trained FP vision transformer $\mathcal{P}$.}
	
	\KwOut{A quantized vision transformer $\mathcal{Q}$.}
	
	Initialize the quantized model $\mathcal{Q}$ by Eq. \ref{eq:quant};
	
	Randomly produce Gaussian noise $\mathcal{G}\thicksim \mathcal{N}(0,1)$;
	
	\For{number of iterations}{

	\texttt{\textbf{\textcolor{darkgray}{\# Stage 1: Sample generation}}}
	{\color[RGB]{218,128,128}$$\mathcal{G} = \arg\max_{\mathcal{G}} \left[\sum_{l=1}^{L}\mathbb{E}(\bm{\Gamma}_l;\mathcal{G})+\alpha\mathcal{D}(\mathcal{O}_\mathcal{Q},\mathcal{O}_\mathcal{P};\mathcal{G})\right]$$}
	
	\vspace{-0.5em}
	\For{g\_step}{
		
		Input $\mathcal{G}$ into $\mathcal{P}$ and $\mathcal{Q}$ to obtain $\mathcal{O}_\mathcal{P}$ and $\mathcal{O}_\mathcal{Q}$;
		
		Hook attention of $\mathcal{P}$ and calculate $\bm{\Gamma}_l$ by Eq. \ref{eq:gamma};
		
		Calculate $\mathcal{L}_{PSE}$ by Eq. \ref{eq:l_pse};
	
		Calculate $\mathcal{L}_{D}$ by Eq. \ref{eq:l_d};

		Combine the losses to obtain $\mathcal{L}_{G}$ by Eq. \ref{eq:l_g};
		
		Fix $\mathcal{P}$ and $\mathcal{Q}$, update $\mathcal{G}$ by minimizing $\mathcal{L}_{G}$;

	}

	\texttt{\textbf{\textcolor{darkgray}{\# Stage 2: Quantization learning}}}
	{\color[RGB]{112,173,71}$$\mathcal{Q} = \arg\min_{\mathcal{Q}} \mathcal{D}(\mathcal{O}_\mathcal{Q},\mathcal{O}_\mathcal{P};\mathcal{G})$$}

	\vspace{-1em}
	\For{q\_step}{

		Perform image augmentation $\hat{\mathcal{G}}$ = aug$(\mathcal{G})$;

		Input $\hat{\mathcal{G}}$ into $\mathcal{P}$ and $\mathcal{Q}$ to obtain $\mathcal{O}_\mathcal{P}$ and $\mathcal{O}_\mathcal{Q}$;

		Calibrate clipping values $q_0$ and $q_{2^k-1}$ of $\mathcal{Q}$;

		Calculate $\mathcal{L}_\mathcal{Q}$ by Eq. \ref{eq:l_q};
		
		Fix $\mathcal{P}$ and $\mathcal{G}$, update $\mathcal{Q}$ by minimizing $\mathcal{L}_{Q}$;
	}

	}
	
\end{algorithm}

We fuse the patch similarity metric and the adaptive teacher-student strategy to obtain the overall scheme, which is summarized in Algorithm 1. Following the teacher-student strategy, the overall pipeline is also performed in a two-stage cycle consisting of sample generation and quantization learning. The procedure and loss functions of each stage are described individually below.

\textbf{Sample generation} In this stage, both the diversity of patch similarity and the model discrepancy are considered to search for a satisfactory sample space. Specifically, we combine the Patch Similarity Entropy Loss $\mathcal{L}_{PSE}$ and the Discrepancy Loss $\mathcal{L}_\mathcal{D}$, and since they are to be maximized, the loss $\mathcal{L}_\mathcal{G}$ of sample generation is defined as follows:
\begin{equation}
	\begin{aligned}
		\mathcal{L}_\mathcal{G} &= - \mathcal{L}_{PSE}-\alpha\cdot \mathcal{L}_\mathcal{D} \\
		& = -\sum_{l=1}^{L}\mathbb{E}(\bm{\Gamma}_l;\mathcal{G}) - \alpha\cdot\left[\frac{1}{n}||\mathcal{Q}(\mathcal{G})-\mathcal{P}(\mathcal{G})||_1 \right] 
	\end{aligned}
	\label{eq:l_g}
\end{equation}
where $\alpha$ is a balance coefficient. Note that in contrast to PSAQ-ViT, there is no auxiliary category prior in the above equation, which ensures the generality of the scheme.

\textbf{Quantization learning} In this stage, we utilize the samples generated in the former stage to facilitate the learning of the quantized model under the supervision of the FP model. However, since the batch size of the generated samples is small (\eg, 32), the fearful overfitting cases typically appear in quantization learning, which poses a challenge to the constant and effective learning of the quantized model.
A naive approach is to simply increase the batch size in the former stages, but this brings unexpected computational overhead. Therefore, we employ image augmentation to cost-effectively expand the sample capacity, following the method in SimCLR \cite{chen2020simple} whose augmentation policy includes random crop (with flip and resize), color distortion, and Gaussian blur.

Then, we input the augmented samples $\hat{\mathcal{G}}$ into the FP model and the quantized model, calibrate the quantization clipping values $q_0$ and $q_{2^k-1}$, and update the quantized model by minimizing the model discrepancy via the following loss:
\begin{equation}
		\mathcal{L}_\mathcal{Q} = \mathcal{L}_{D} = \frac{1}{n}||\mathcal{Q}(\hat{\mathcal{G}})-\mathcal{P}(\hat{\mathcal{G}})||_1
	\label{eq:l_q}
\end{equation}

After several iterations of learning in the above two stages, the generated samples and the quantized model will reach an ideal balance point, where the model discrepancy no longer changes after iterations, \ie, the generated samples cannot pull apart the FP model and the quantized model, and the quantized model have learned the full transferred knowledge. In this case, the quantized model is considered to be functionally identical to the FP model. 

It is also worth noting that all the objective functions are task-independent, thus the proposed scheme is general-purpose and compatible with a broad range of vision tasks, and they are also model-independent, which allows for different teacher and student model structures, \eg, the learning of quantized DeiT-T can be performed under the supervision of DeiT-B.

\section{Experiments}

\subsection{Implementation Details}

\textbf{Models and datasets} We evaluate PSAQ-ViT V2 on various popular models for image classification, object detection, and semantic segmentation with the following models and datasets.    
\begin{itemize}
	\item Image classification: We perform classification experiments on ImageNet \cite{deng2009imagenet} dataset, which has 1.28M training images and 50K validation images from 1,000 classes. DeiT-T/S/B \cite{touvron2021training} and Swin-T/B \cite{liu2021swin} are adopted as the baseline model. Since ViT \cite{dosovitskiy2020image} and DeiT have the same model structure, we do not evaluate ViT in this work. The pre-trained models are obtained from the timm \cite{rw2019timm} framework.
	\item Object detection: Object detection and instance segmentation experiments are conducted on COCO 2017 \cite{lin2014microsoft} dataset, which contains 118K training, 5K validation, and 20K test-dev images. The baseline is the Cascade Mask R-CNN \cite{cai2018cascade} framework in mmdetection \cite{mmdetection} with DeiT-T/S and Swin-T as the backbones.
	\item Semantic segmentation: We adopt ADE20K \cite{zhou2019semantic} dataset to evaluate the performance for semantic segmentation. It has 25K images in total, with 20K for training, 2K for validation, and another 3K for testing, covering a broad range of 150 semantic categories. With DeiT-T/S and Swin-T as the backbones, we take the UperNet \cite{xiao2018unified} framework in mmsegmentation \cite{mmseg2020} as the baseline.
\end{itemize}

\textbf{Comparison methods} It is naturally necessary to make a comparison with the previous version PSAQ-ViT, however, it is unfair and inadequate to compare with only PSAQ-ViT, because PSAQ-ViT only performs quantization parameter calibration without a learning process, and it contains the category prior, which is not applicable to detection and segmentation tasks. Therefore, {\color{black} we have to build a reasonable comparison method, called \emph{Standard V2}, on our own. 
Standard V2 is distinguished from Standard in the previous version.
Standard V2 introduces the model interaction} and directly uses real training data with the Discrepancy Loss $\mathcal{L}_\mathcal{D}$ for teacher-student learning, and it traverses the training set once to complete convergence. For a fair comparison, we make sure to use the same number of samples, more intuitively, in PSAQ-ViT V2, the result of multiplying the number of iterations, q\_step, and batch size is equal to the size of the training set. All other settings for {\color{black}Standard V2} and PSAQ-ViT V2 are the same.

\textbf{Experimental settings} All experiments in this work are implemented in Pytorch. For a fair demonstration of effectiveness, we employ the most basic quantization method, \ie, for weights, symmetric uniform quantization with vanilla MinMax strategy is applied; for activations, asymmetric uniform quantization is applied, and the default strategy is vanilla MinMax if not specifically declared.
We adopt Adam \cite{kingma2014adam} optimizer in both sample generation and quantization learning stages, where the search space of learning rate is \{0.2, 0.25\} in sample generation and \{5e-7, 1e-6, 2e-6\} in quantization learning. {\color{black}Although further hyperparameter tuning may achieve better accuracy, for uniformity, all our experiments are conducted using batch size 32 and weight decay 1e-4.}
The hyperparameter $\alpha$ is set to 1.0 after a simple grid search, and its selection has little effect on the final performance. {\color{black} All experiments are done on a single NVIDIA GeForce RTX 3090 GPU.}

\subsection{Quantization Results for Image Classification}
We start by discussing the quantization results on ImageNet image classification, which is compared with the previous version PSAQ-ViT and the real-data-driven {\color{black}Standard V2}. 
Interestingly, our proposed PSAQ-ViT V2 without access to the original dataset can even slightly \emph{outperforms} {\color{black}Standard V2}. This is partly thanks to the Patch Similarity Entropy Loss $\mathcal{L}_{PSE}$. It forces the distinction between foreground and background in the generated samples according to the response of the self-attention module, and then when these samples are utilized to calibrate the quantization parameters, they in turn reinforce the functionality of the self-attention module, thus acting as positive feedback that can reduce the activation outliers to some extent and therefore improve the tolerance to parameter clipping.
The Discrepancy Loss $\mathcal{L}_\mathcal{D}$ also contributes to the surprising performance. It encourages increasing the discrepancy between the FP model and the quantized model to force the emergence of new hard-to-discriminate features in the samples, and these features can be regarded as support vectors (aka hard instances) that contain more typical and instructive knowledge, thus promoting more effective learning of the quantized model.

\begin{table}[t]\scriptsize
	\begin{center}
		\caption{Quantization results on ImageNet Image classification. We compare the proposed PSAQ-ViT V2 with the previous version PSAQ-ViT, as well as {\color{black}Standard V2}, which uses an equal number of real samples for direct teacher-student learning.
		PSAQ-ViT V2 significantly improves the quantization accuracy, even slightly outperforming real-data-driven {\color{black}Standard V2}. Here, ``No Data" indicates that no real data participate in the quantization process, and ``W$x$/A$y$" denotes quantifying the weights and activations to $x$-bit and $y$-bit, respectively. {\color{black} We abbreviate quantization precision as ``Prec", model size as ``Size" (in MB), Bit Operations \cite{van2020bayesian} as ``BOPS" (in G), and Top-1 test accuracy as ``Top-1" (in \%).}}
		\label{exp:ImageNet}
		\begin{tabular}{ccccccc}
			\toprule
			Model & Method & No Data & Prec. & Size & {\color{black}BOPS} & Top-1 \\
			
			\midrule
			\multirow{8.35}*{\tabincell{c}{DeiT-T}} 
			& Baseline & $-$ & FP & 20 & {\color{black}1106} & 72.21  \\
			\cmidrule{2-7}
			& PSAQ-ViT & $\checkmark$ & W4/A8 & 2.5 & {\color{black}34.6} & 65.57 \\
			& {\color{black}Standard V2} & $\times$ & W4/A8 & 2.5 & {\color{black}34.6} &  68.43 \\
			& PSAQ-ViT V2 & $\checkmark$ & W4/A8 & 2.5 &  {\color{black}34.6} & \textbf{68.61} \\
			\cmidrule{2-7}
			& PSAQ-ViT & $\checkmark$ & W8/A8 & 5 & {\color{black}69.1} & 71.56 \\
			& {\color{black}Standard V2} & $\times$ & W8/A8 & 5 & {\color{black}69.1} &  72.06 \\
			& PSAQ-ViT V2 & $\checkmark$ & W8/A8 & 5 & {\color{black}69.1} & \textbf{72.17} \\

			\midrule
			\multirow{8.35}*{\tabincell{c}{DeiT-S}} 
			& Baseline & $-$ & FP & 88 & {\color{black}4710} & 79.85  \\
			\cmidrule{2-7}
			& PSAQ-ViT & $\checkmark$ & W4/A8 & 11 & {\color{black}147} & 73.23 \\
			& {\color{black}Standard V2} & $\times$ & W4/A8 & 11 & {\color{black}147} & 75.98  \\
			& PSAQ-ViT V2 & $\checkmark$ & W4/A8 & 11 & {\color{black}147} & \textbf{76.36} \\
			\cmidrule{2-7}
			& PSAQ-ViT & $\checkmark$ & W8/A8 & 22 & {\color{black}294} & 76.92 \\
			& {\color{black}Standard V2} & $\times$ & W8/A8 & 22 & {\color{black}294} & 79.24  \\
			& PSAQ-ViT V2 & $\checkmark$ & W8/A8 & 22 & {\color{black}294} & \textbf{79.56} \\

			\midrule
			\multirow{8.35}*{\tabincell{c}{DeiT-B}} 
			& Baseline & $-$ & FP & 344 & {\color{black}17920} & 81.85  \\
			\cmidrule{2-7}
			& PSAQ-ViT & $\checkmark$ & W4/A8 & 43 & {\color{black}560} & 77.05 \\
			& {\color{black}Standard V2} & $\times$ & W4/A8 & 43 & {\color{black}560} &  79.17 \\
			& PSAQ-ViT V2 & $\checkmark$ & W4/A8 & 43 & {\color{black}560} & \textbf{79.49} \\
			\cmidrule{2-7}
			& PSAQ-ViT & $\checkmark$ & W8/A8 & 86 & {\color{black}1120} & 79.10 \\
			& {\color{black}Standard V2} & $\times$ & W8/A8 & 86 & {\color{black}1120} &  81.26 \\
			& PSAQ-ViT V2 & $\checkmark$ & W8/A8 & 86 & {\color{black}1120} & \textbf{81.52} \\

			\midrule
			\multirow{8.35}*{\tabincell{c}{Swin-T}} 
			& Baseline & $-$ & FP & 116 & {\color{black}4608} & 81.35  \\
			\cmidrule{2-7}
			& PSAQ-ViT & $\checkmark$ & W4/A8 & 14.5 & {\color{black}144} & 71.79 \\
			& {\color{black}Standard V2} & $\times$ & W4/A8 & 14.5 & {\color{black}144} &  75.51 \\
			& PSAQ-ViT V2 & $\checkmark$ & W4/A8 & 14.5 & {\color{black}144} & \textbf{76.28} \\
			\cmidrule{2-7}
			& PSAQ-ViT & $\checkmark$ & W8/A8 & 29 & {\color{black}288} & 75.35 \\
			& {\color{black}Standard V2} & $\times$ & W8/A8 & 29 & {\color{black}288} & 79.62  \\
			& PSAQ-ViT V2 & $\checkmark$ & W8/A8 & 29 & {\color{black}288} & \textbf{80.21} \\

			\midrule
			\multirow{8.35}*{\tabincell{c}{Swin-S}} 
			& Baseline & $-$ & FP & 200 & {\color{black}8909} & 83.20  \\
			\cmidrule{2-7}
			& PSAQ-ViT & $\checkmark$ & W4/A8 & 25 & {\color{black}278} & 75.14 \\
			& {\color{black}Standard V2} & $\times$ & W4/A8 & 25 & {\color{black}278} &  78.22 \\
			& PSAQ-ViT V2 & $\checkmark$ & W4/A8 & 25 & {\color{black}278} & \textbf{78.86} \\
			\cmidrule{2-7}
			& PSAQ-ViT & $\checkmark$ & W8/A8 & 50 & {\color{black}557} & 76.64 \\
			& {\color{black}Standard V2} & $\times$ & W8/A8 & 50 & {\color{black}557} &  81.42 \\
			& PSAQ-ViT V2 & $\checkmark$ & W8/A8 & 50 & {\color{black}557} & \textbf{82.13} \\
			
			\bottomrule
		\end{tabular}
	\end{center}
\end{table}

Table \ref{exp:ImageNet} shows the quantization results for various models at different quantization bit-precisions.
For the quantization of DeiT-T, our proposed PSAQ-ViT V2 improves 3.04\% over the previous version at W4/A8 precision, and achieves 72.17\% accuracy at W8/A8 precision with only 5MB memory footprint.
DeiT-S quantized with our method improves 0.38\% and 0.32\% over the real-data-driven {\color{black}Standard V2} at W4/A8 and W8/A8 precisions, respectively.
Similar to the previous models, the quantization results of DeiT-B also show that PSAQ-ViT V2 achieves the best performance, with 2.44\% and 0.32\% improvements over PSAQ-ViT and {\color{black}Standard V2} at W4/A8 precision, respectively, and only 0.33\% accuracy reduction compared to the FP model at W8/A8 precision with 4-fold compression.
When using the vanilla MinMax quantization strategy, the parameter distribution of Swin is highly sensitive to discretization; for instance, PSAQ-ViT produces 9.56\% and 6.00\% accuracy degradation in the W4/A8 and W8/A8 quantization of Swin-T, respectively.
Fortunately, the proposed teacher-student strategy updates the parameter distribution in the model to make it quantization-friendly, and thus achieves significant accuracy compensation of 4.49\% and 4.86\% at W4/A8 and W8/A8 precisions, respectively.
The adaptive nature of our method is also more evident in Swin; for instance, for the W4/A8 and W8/A8 quantization of Swin-S, our method achieves performance improvements of 0.64\% and 0.71\% over {\color{black}Standard V2}, respectively.

\textbf{Results of combining with advanced strategies}
Since our method is orthogonal to the quantization strategies, we combine PSAQ-ViT V2 with advanced quantization strategies to further improve the performance, as shown in Table \ref{exp:advanced}.
First, we employ exponential moving average (EMA), which smooths the quantization clipping values, to replace vanilla MinMax, achieving an accuracy gain of 0.97\% and 1.05\% in the W4/A8 quantization of DeiT-T and Swin-T, respectively.
Moreover, as mentioned before, the model prior and loss functions in PSAQ-ViT V2 are model-independent, thus allowing for different teacher and student model structures. Here, we employ the base model to supervise the quantization of the corresponding tiny model.
For instance, thanks to the guidance of the powerful FP DeiT-B with 81.85\%, the quantization performance of DeiT-T at W4/A8 precision achieves an inspiring 1.25\% improvement.

\setlength{\tabcolsep}{3.8pt}
\begin{table}[t]\scriptsize
	\begin{center}
		\caption{Quantization results of combining with advanced strategies, including EMA for smoothing the quantization clipping values and supervised learning by a better teacher. For the latter, here we employ the base model to supervise the quantization of the corresponding tiny model.}
		\label{exp:advanced}
		\begin{tabular}{cccccccc}
			\toprule
			Model & Method & Advanced & No Data & Prec. & Size & {\color{black}BOPS} & Top-1 \\

			\midrule
			\multirow{4.35}*{\tabincell{c}{DeiT-T}} 
			& Baseline & $-$ & $-$ & FP & 20 & {\color{black}1106} & 72.21  \\
			\cmidrule{2-8}
			& \multirow{3}*{PSAQ-ViT V2}   & $-$ & $\checkmark$ & W4/A8 & 2.5 & {\color{black}34.6} & 68.61 \\
			& & EMA & $\checkmark$ & W4/A8 & 2.5 & {\color{black}34.6} & \textbf{69.58} \\
			& & $\mathcal{P}$:DeiT-B & $\checkmark$ & W4/A8 & 2.5 & {\color{black}34.6} & \textbf{69.86} \\

			\midrule
			\multirow{4.35}*{\tabincell{c}{Swin-T}} 
			& Baseline & $-$ & $-$ & FP & 116 & {\color{black}4608} & 81.35  \\
			\cmidrule{2-8}
			& \multirow{3}*{PSAQ-ViT V2}   & $-$ & $\checkmark$ & W4/A8 & 14.5 & {\color{black}144} & 76.28 \\
			& & EMA & $\checkmark$ & W4/A8 & 14.5 & {\color{black}144} & \textbf{77.33} \\
			& & $\mathcal{P}$:Swin-B & $\checkmark$ & W4/A8 & 14.5 & {\color{black}144} & \textbf{77.19} \\
			
			\bottomrule
		\end{tabular}
	\end{center}
\end{table}

\begin{figure}[t]
	\centering
	\subfloat[Noise]{
	  \includegraphics[width=0.112\linewidth]{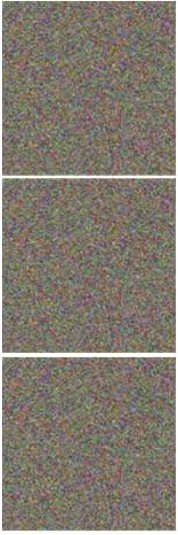}
	}
	\subfloat[PSAQ-ViT]{
	  \includegraphics[width=0.224\linewidth]{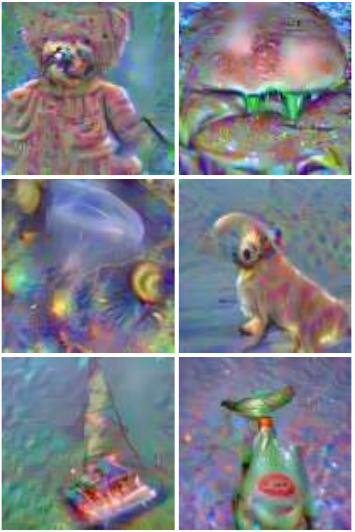}
	}
	\subfloat[PSAQ-ViT V2 (Mid iteration) \centering]{
	  \includegraphics[width=0.224\linewidth]{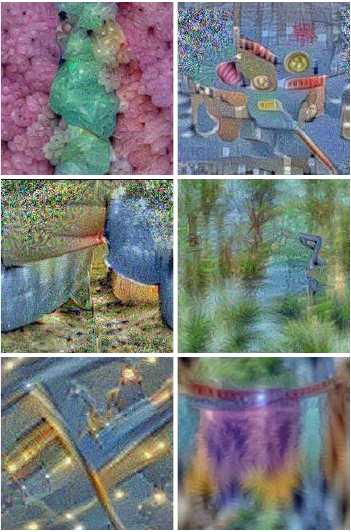}
	}
	\subfloat[PSAQ-ViT V2 \\(Final iteration) \centering]{
	  \includegraphics[width=0.336\linewidth]{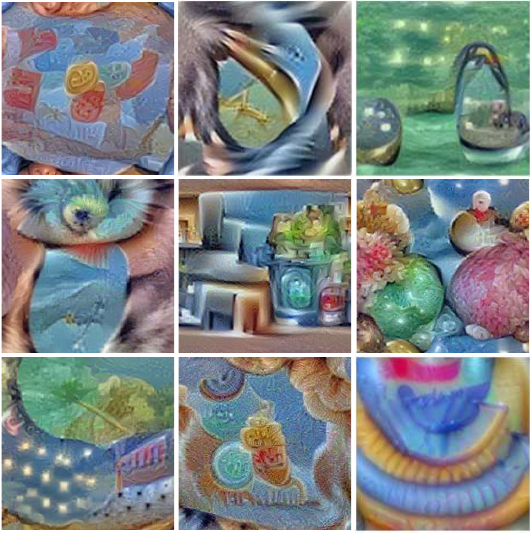}
	}
	\caption{The fake samples (224$\times$224 pixels) generated using PSAQ-ViT and PSAQ-ViT V2 (including mid and final iterations), which only need a pre-trained FP model on ImageNet and no additional information. Compared to the previous version with category guidance, the samples generated by this work are more informative and diverse. In addition, during the cyclic iterations, the samples are progressively evolving, with continuously emerging new features and increasingly evident semantic information, thus facilitating constant and effective learning of the quantized model.}
	\label{fig:vis}
\end{figure}

\textbf{Analysis of generated samples}
Fig. \ref{fig:vis} shows the visualization results of the generated images (224$\times$224 pixels) using PSAQ-ViT and PSAQ-ViT V2 (including mid and final iterations). Note that these images require only a pre-trained FP model, and not any additional information, especially the original data or any absolute value metrics. Thanks to the joint guidance of the patch similar metric and the category prior, the generated images in PSAQ-ViT can clearly distinguish the foreground from the background, and the foreground is rich in semantic information with a specified category, \eg, the top left pseudo-label is ``teddy bear". However, the background information in these samples is singular and homogeneous.
In contrast, there is no category guidance in this work, resulting in weakly intuitive semantic features of the images, but this in turn helps the sample to improve informativeness and diversity in both foreground and background without limitations, thus more closely matching the real data in the overall statistics.
In addition, during the cyclic iterations, the samples are progressively evolving in competition and interaction, with continuously emerging new features and increasingly evident semantic information, thus facilitating constant and effective learning of the quantized model.

\begin{figure}[t]
	\centering
	\includegraphics[width=1.0\linewidth]{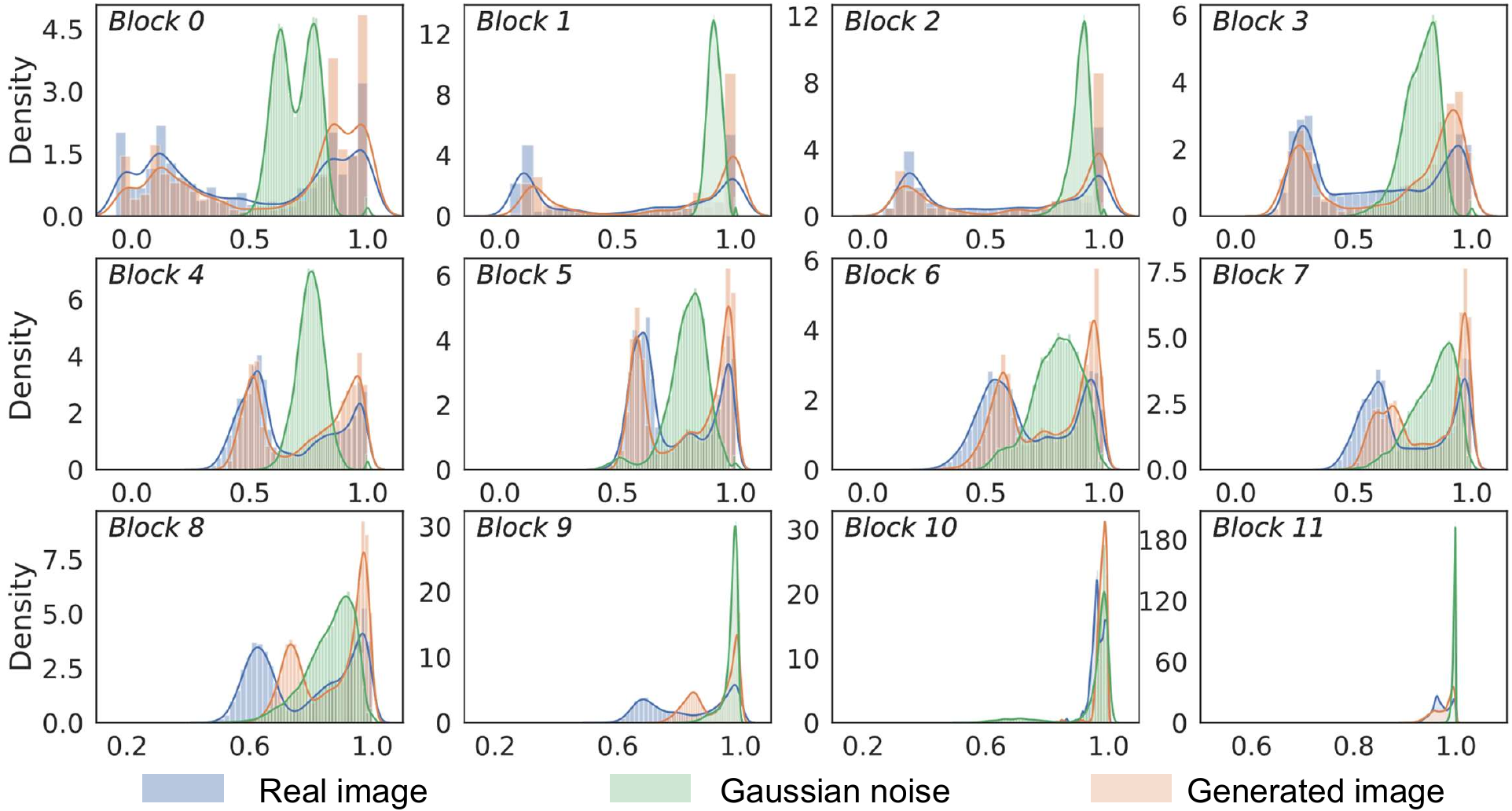}
	\caption{Comparison of the kernel density curves of the patch similarity for each block in DeiT-B model when the input is the real image, Gaussian noise, and the generated image. The x-axis represents the values of patch similarity. As we can see, the density of each block corresponding to Gaussian noise shows a concentrated unimodal shape, while the generated image and the real image have similar properties, producing the density with a dispersed bimodal shape.}
	\label{fig:compare12}
\end{figure}

To demonstrate the validity of the Patch Similarity Entropy Loss $\mathcal{L}_{PSE}$ more obviously, we also conduct the comparison experiments of the kernel density curves of the patch similarity for each block in DeiT-B model when the input is the real image, Gaussian noise, and the generated image, as shown in Fig. \ref{fig:compare12}. For the responses to Gaussian noise, the kernel density curves all show a concentrated unimodal shape and the central value of the curve is high, indicating a high degree of similarity between each patch of Gaussian noise and thus a full classification as background or foreground.
Fortunately, the kernel density curves corresponding to our generated images are very approximate to those corresponding to the real images. They all show a dispersed bimodal shape, indicating a high diversity of responses, and the left and right peaks of curves describe inter- and intra-category similarity, respectively, which is in line with the expectation that the images can easily be distinguished between foreground and background.

\subsection{Quantization Results for Object Detection}
Unlike PSAQ-ViT which is only applicable to the image classification task, PSAQ-ViT V2 is general and compatible with a broad range of vision tasks. Here, the Cascade Mask R-CNN framework with Swin-S/DeiT-S as the backbone, which achieves state-of-the-art (SOTA) performance on object detection and instance segmentation tasks on COCO dataset, is used to perform quantization experiments and the results are reported in Table \ref{exp:COCO}.
To the best of our knowledge, this is the first attempt to perform data-free quantization of ViTs for object detection and instance segmentation.

When using DeiT-S as the backbone, PSAQ-ViT V2 achieves consistently better performance than {\color{black}Standard V2}, with 0.3 higher box {\color{black}Average Precision (AP)} and 0.4 higher mask AP at W4/A8 precision, and it achieves 47.3 box AP and 40.8 mask AP at W8/A8 precision, enabling a 4-fold almost lossless compression.
In the case of applying Swin-S as the backbone, our method also shows promising performance, outperforming {\color{black}Standard V2} by 0.6 box AP and 0.6 mask AP at W8/A8 precision, which is only 0.9 box AP and 0.6 mask AP lower than the FP model.

\setlength{\tabcolsep}{2.2pt}
\begin{table}[t]\scriptsize
	\begin{center}
		\caption{Quantization results on COCO object detection and instance segmentation. We take the concatenation of the backbone DeiT-S/Swin-S and the detection method Cascade Mask R-CNN as the baseline models. PSAQ-ViT V2 can even slightly outperforms real-data-driven {\color{black}Standard V2}. Here, ``AP$^{\text{box}}$" is the box AP for object detection, and ``AP$^{\text{mask}}$" is the mask AP for instance segmentation.}
		\label{exp:COCO}
		\begin{tabular}{cccccccc}
			\toprule
			Model & Method & No Data & Prec. & Size & {\color{black}BOPS($\times 10^3$)} & AP$^{\text{box}}$ & AP$^{\text{mask}}$ \\
			
			\midrule
			\multirow{6.3}*{\tabincell{c}{DeiT-S / \\ Cascade \\ Mask R-CNN}} 
			& Baseline & $-$ & FP & 320 & {\color{black}910} & 48.0 & 41.4 \\
			\cmidrule{2-8}
			& {\color{black}Standard V2} & $\times$ & W4/A8 & 40 & {\color{black}28.4} & 44.5 & 38.4 \\
			& PSAQ-ViT V2 & $\checkmark$ & W4/A8 & 40 & {\color{black}28.4} & \textbf{44.8} & \textbf{38.8} \\
			\cmidrule{2-8}
			& {\color{black}Standard V2} & $\times$ & W8/A8 & 80 & {\color{black}56.9} & 46.8 & 40.2 \\
			& PSAQ-ViT V2 & $\checkmark$ & W8/A8 & 80 & {\color{black}56.9} & \textbf{47.3} & \textbf{40.8} \\

			\midrule
			\multirow{6.3}*{\tabincell{c}{Swin-S / \\ Cascade \\ Mask R-CNN}} 
			& Baseline & $-$ & FP & 428 & {\color{black}858} & 51.8 & 44.7 \\
			\cmidrule{2-8}
			& {\color{black}Standard V2} & $\times$ & W4/A8 & 50.4 & {\color{black}26.8} & 47.2 & 40.8 \\
			& PSAQ-ViT V2 & $\checkmark$ & W4/A8 & 50.4 & {\color{black}26.8} & \textbf{47.9} & \textbf{41.4} \\
			\cmidrule{2-8}
			& {\color{black}Standard V2} & $\times$ & W8/A8 & 107 & {\color{black}53.6} & 50.3 &  43.5 \\
			& PSAQ-ViT V2 & $\checkmark$ & W8/A8 & 107 & {\color{black}53.6} & \textbf{50.9} & \textbf{44.1} \\
			
			\bottomrule
		\end{tabular}
	\end{center}
\end{table}

\begin{figure*}[t]
	\centering
	\includegraphics[width=0.9\linewidth]{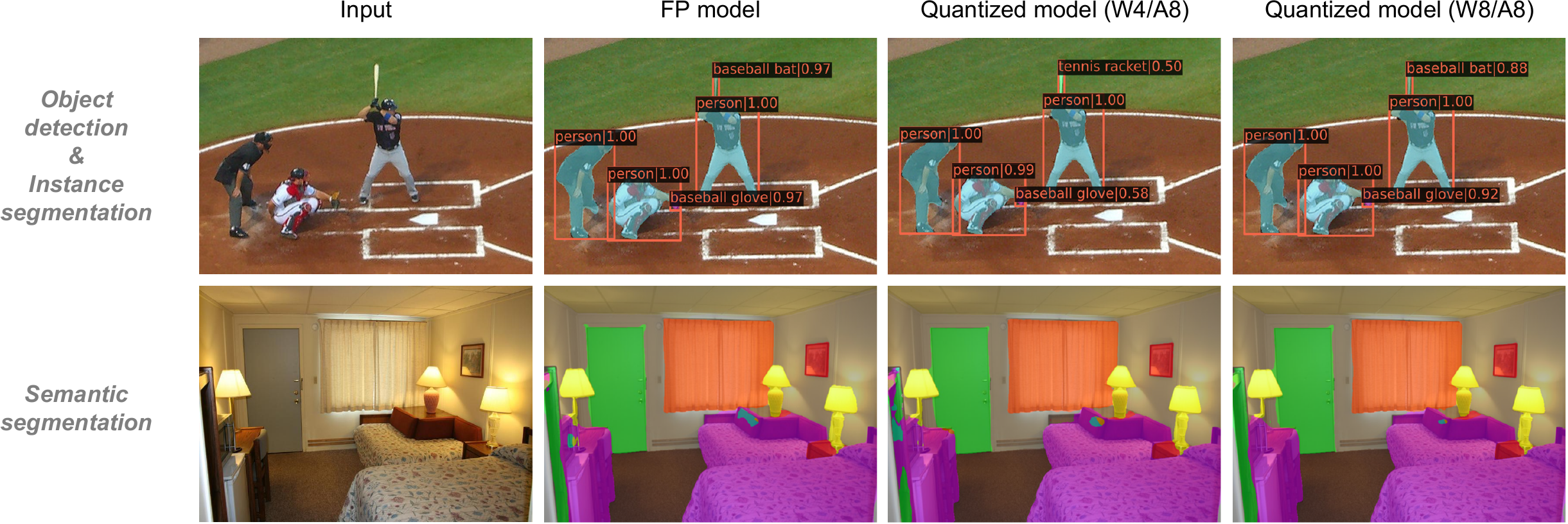}
	\caption{Visualization of detection and segmentation results for the FP model and the quantized model. The backbone is Swin-S. The first row shows the results of object detection and instance segmentation on COCO dataset, and the second row shows the results of semantic segmentation on ADE20K dataset. The performance of the quantized model with W8A8 precision is comparable to that of the FP model, and there is a slight performance degradation with W4A8 precision, for instance, in object detection, the results of easy objects such as ``person" are unaffected, but the hard small object ``baseball bat" is misclassified into ``tennis racket".}
	\label{fig:vis_det_seg}
\end{figure*}

For a more intuitive view of the performance of the quantized model, we visualize the detection results and compare them with those of the FP model, as illustrated in the first row in Fig. \ref{fig:vis_det_seg}.
The backbone we employ is Swin-S. 
It can be seen that the quantized model at W8/A8 precision achieves similar performance to the FP model, with slight differences in partial classification scores, \eg, the score for 'baseball glove' changes from 0.97 to 0.92. And when performing W4/A8 quantization, the hard small object ``baseball bat" is misclassified into ``tennis racket", while the results of easy objects such as ``person" are unaffected.

\subsection{Quantization Results for Semantic Segmentation}
Our proposed PSAQ-ViT V2 can also be applied to the semantic segmentation task. We employ the UperNet framework with Swin-S/DeiT-S as the backbone, which is the SOTA scheme on the semantic segmentation task on ADE20K dataset, and the quantization results are shown in Table \ref{exp:ADE20K}.
To the best of our knowledge, this is the first attempt to quantize ViTs for semantic segmentation without any real data.

The discretization of the model parameters has a large impact on the pixel-level semantic segmentation, making the accuracy of the quantized model not as encouraging as the results on previous tasks. For instance, W8/A8 quantization cannot achieve lossless compression, but rather produces an accuracy degradation greater than 1 {\color{black} mean Intersection over Union (mIoU)}.
This effect also acts on {\color{black}Standard V2} using real data, thus the performance advantage of our method over {\color{black}Standard V2} does not change.
For the quantization of DeiT-S/UperNet, our proposed PSAQ-ViT V2 achieves 39.9 mIoU and 42.7 mIoU at W4/A8 and W8/A8 precisions, respectively, which are 0.3 mIoU and 0.2 mIoU higher than {\color{black}Standard V2}.
Similarly, in the W4/A8 and W8/A8 quantization of Swin-S/UperNet, our method also consistently has performance benefits over {\color{black}Standard V2}.

In addition, we also perform visualizations of the segmentation results, as illustrated in the second row in Fig. \ref{fig:vis_det_seg}.
The backbone we employ is Swin-S. 
As we can see, the quantized model at W8/A8 precision is comparable to the FP model for the overall segmentation performance, while there is a slight difference in the recognition of a few small regions.
In the case of compressing the model to W4/A8 precision, the segmentation results are not good for semantically weak regions, for instance, the object in the lower left corner of the image is not well recognized.

\setlength{\tabcolsep}{4.5pt}
\begin{table}[t]\scriptsize
	\begin{center}
		\caption{Quantization results on ADE20K semantic segmentation. We take the concatenation of the backbone DeiT-S/Swin-S and the segmentation method UperNet as the baseline models. PSAQ-ViT V2 can even slightly outperforms real-data-driven {\color{black}Standard V2}.}
		\label{exp:ADE20K}
		\begin{tabular}{ccccccc}
			\toprule
			Model & Method & No Data & Prec. & Size & {\color{black} BOPS($\times 10^3$)} & mIoU  \\
			
			\midrule
			\multirow{6.3}*{\tabincell{c}{DeiT-S / \\ UperNet}} 
			& Baseline & $-$ & FP & 208 & {\color{black} 1125} & 44.0 \\
			\cmidrule{2-7}
			& {\color{black}Standard V2} & $\times$ & W4/A8 & 26 & {\color{black} 35.2} & 39.6 \\
			& PSAQ-ViT V2 & $\checkmark$ & W4/A8 & 26 & {\color{black} 35.2} & \textbf{39.9} \\
			\cmidrule{2-7}
			& {\color{black}Standard V2} & $\times$ & W8/A8 & 52 & {\color{black} 70.3} & 42.5 \\
			& PSAQ-ViT V2 & $\checkmark$ & W8/A8 & 52 & {\color{black} 70.3} & \textbf{42.7} \\

			\midrule
			\multirow{6.3}*{\tabincell{c}{Swin-S / \\ UperNet}} 
			& Baseline & $-$ & FP & 324 & {\color{black} 1063} & 49.3  \\
			\cmidrule{2-7}
			& {\color{black}Standard V2} & $\times$ & W4/A8 & 40.5 & {\color{black} 33.2} &  44.2 \\
			& PSAQ-ViT V2 & $\checkmark$ & W4/A8 & 40.5 & {\color{black} 33.2} & \textbf{44.6} \\
			\cmidrule{2-7}
			& {\color{black}Standard V2} & $\times$ & W8/A8 & 81 & {\color{black} 66.4} &  46.7 \\
			& PSAQ-ViT V2 & $\checkmark$ & W8/A8 & 81 & {\color{black} 66.4} & \textbf{47.2} \\
			
			\bottomrule
		\end{tabular}
	\end{center}
\end{table}

\subsection{Ablation Studies}
Here, we perform two ablation studies to verify the effectiveness of each component of the proposed method. First, we explore the contributions of the Patch Similarity Entropy Loss $\mathcal{L}_{PSE}$ and the Discrepancy Loss $\mathcal{L}_{\mathcal{D}}$ in sample generation, as shown in Table \ref{exp:ablation}.
Without any optimization, inputting Gaussian noise directly to guide the learning of the quantized model can produce severe performance degradation.
When only $\mathcal{L}_{\mathcal{D}}$ is utilized on Gaussian noise to perform the minimax game, there is a slight improvement in accuracy, but it is still far from satisfactory.
{\color{black} In summary, in the absence of real data, the quantization performance obtained directly with Gaussian noise is unacceptable. Thus, $\mathcal{L}_{PSE}$-guided optimization to make the Gaussian noise approximate the real images is highly necessary. For instance, when performing W4/A8 quantization of DeiT-S, only $\mathcal{L}_{PSE}$-guided optimization achieves 74.69\% accuracy, which is 59.97\% higher than Gaussian noise. Further, we add $\mathcal{L}_{\mathcal{D}}$ to introduce the model interaction so that the quantized model enjoys a more adequate knowledge transfer, resulting in a better performance with a 1.67\% improvement. Similar comparative results are also obtained in the W4/A8 quantization of Swin-S, where using $\mathcal{L}_{PSE}$ improves the accuracy substantially to 77.03\% and combining it with $\mathcal{L}_{\mathcal{D}}$ further improves to a satisfactory 78.86\%. The above results fully demonstrate the critical role of $\mathcal{L}_{PSE}$ and $\mathcal{L}_{\mathcal{D}}$ in improving the quantization performance.} 

\setlength{\tabcolsep}{7.5pt}
\begin{table}[t]\scriptsize
	\begin{center}
		\caption{Ablation studies on the losses for updating the generated samples $\mathcal{G}$. The Patch Similarity Entropy Loss $\mathcal{L}_{PSE}$ plays a crucial role in sample generation. On this basis, the model interaction introduced by the Discrepancy Loss $\mathcal{L}_{\mathcal{D}}$ also substantially improves the quantization accuracy.}
		\label{exp:ablation}
		\begin{tabular}{ccccccc}
			\toprule
			Model & $\mathcal{L}_{PSE}$ & $\mathcal{L}_{D}$ & Prec. & Size & {\color{black} BOPS} & Top-1 \\

			\midrule
			\multirow{5.5}*{\tabincell{c}{DeiT-S}} 
			& $-$ & $-$ & FP & 88 & {\color{black} 4710} & 79.85 \\
			\cmidrule{2-7}
			& $\times$ & $\times$ & W4/A8 & 11 & {\color{black} 147} & 14.72 \\
			& $\times$ & $\checkmark$ & W4/A8 & 11 & {\color{black} 147} & 25.81 \\
			& $\checkmark$ & $\times$ & W4/A8 & 11 & {\color{black} 147} &  74.69 \\
			& $\checkmark$ & $\checkmark$ & W4/A8 & 11 & {\color{black} 147} & \textbf{76.36} \\

			\midrule
			\multirow{5.5}*{\tabincell{c}{Swin-S}} 
			& $-$ & $-$ & FP & 200 & {\color{black} 8909} & 83.20 \\
			\cmidrule{2-7}
			& $\times$ & $\times$ & W4/A8 & 25 & {\color{black} 278} & 8.96 \\
			& $\times$ & $\checkmark$ & W4/A8 & 25 & {\color{black} 278} & 20.24 \\
			& $\checkmark$ & $\times$ & W4/A8 & 25 & {\color{black} 278} &  77.03 \\
			& $\checkmark$ & $\checkmark$ & W4/A8 & 25 & {\color{black} 278} & \textbf{78.86} \\
			
			\bottomrule
		\end{tabular}
	\end{center}
\end{table}

Second, we compare the performance of different estimation functions for the model discrepancy $\mathcal{D}(\mathcal{O}_\mathcal{Q},\mathcal{O}_\mathcal{P};\mathcal{G})$, and the results are reported in Table \ref{exp:ablation1}. The experimental results verify the previous analysis that KL divergence leads to a dying minimax game since it produces decaying gradients after converging to the current generated samples.
{\color{black} The reason is that in our minimax game, the student greedily performs one-time learning from the teacher, which leaves little room for optimizing KL divergence, so that the KL divergence results are similar when inputting samples with new features, thus hinders the competition and constant evolution of the generated samples and the quantized model. More specifically, we evaluate the performance of KL divergence for multiple temperature settings $T\in$ \{20, 10, 5, 1\} and find that as the temperature preliminarily decreases (\eg, from 20 to 10), the constraint on the soft label is tightened and the above issue is slightly alleviated. Intuitively, in the W4/A8 quantization of DeiT-S, the accuracy is improved from 74.56\% to 75.23\%. Whereas, after the temperature is small, further reduction does not improve the performance any further, for example, similar performance (75.57\% and 75.51\%) is achieved at temperatures of 5 and 1. This indicates that the above issues are not radically ameliorated. In addition, we introduce an advanced dynamic temperature strategy \cite{jafari2021annealing}, which offers some improvements over the scheme with the fixed temperature, however, it is still weaker than the scheme with the MAE function, with the latter being 0.46\% and 0.31\% higher in the W4/A8 quantization of DeiT-S and Swin-S, respectively. The above results fully demonstrate the validity of the MAE function,} which provides a better search space for the minimax game and thus can effectively solve the gradient decay issue.

\setlength{\tabcolsep}{7pt}
\begin{table}[t]\scriptsize
	\begin{center}
		\caption{Ablation Studies on the estimation functions of the model discrepancy $\mathcal{D}(\mathcal{O}_\mathcal{Q},\mathcal{O}_\mathcal{P};\mathcal{G})$. KL divergence produces decaying gradients after converging to the current generated samples, leading to an unsatisfactory minimax game. Here, T is the temperature coefficient. In contrast, the MAE we employ is the promising function in the minimax game.}
		\label{exp:ablation1}
		\begin{tabular}{cccccc}
			\toprule
			Model & $\mathcal{D}(\mathcal{O}_\mathcal{Q},\mathcal{O}_\mathcal{P};\mathcal{G})$ & Prec. & Size & {\color{black} BOPS} & Top-1 \\

			\midrule
			\multirow{7.3}*{\tabincell{c}{DeiT-S}} 
			& $-$ & FP & 88 & {\color{black} 4710} & 79.85 \\
			\cmidrule{2-6}
			& {\color{black}KLD ($T$=20)} & {\color{black}W4/A8} & {\color{black}11} & {\color{black} 147} &  {\color{black}74.56} \\
			& KLD ($T$=10) & W4/A8 & 11 & {\color{black} 147} & 75.23  \\
			& KLD ($T$=5) & W4/A8 & 11 & {\color{black} 147} & 75.57  \\
			& {\color{black}KLD ($T$=1)} & {\color{black}W4/A8} & {\color{black}11} & {\color{black} 147} &  {\color{black}75.51} \\
			& {\color{black}KLD ($T$:dynamic\cite{jafari2021annealing})} & {\color{black}W4/A8} & {\color{black}11} & {\color{black} 147} &  {\color{black}75.90} \\
			& MAE & W4/A8 & 11 & {\color{black} 147} & \textbf{76.36} \\

			\midrule
			\multirow{7.3}*{\tabincell{c}{Swin-S}} 
			& $-$ & FP & 200 & {\color{black} 8909} & 83.20 \\
			\cmidrule{2-6}
			& {\color{black}KLD ($T$=20)} & {\color{black}W4/A8} & {\color{black}25} & {\color{black} 278} & {\color{black}77.02} \\
			& KLD ($T$=10) & W4/A8 & 25 & {\color{black} 278} &  77.67 \\
			& KLD ($T$=5) & W4/A8 & 25 & {\color{black} 278} & 78.04  \\
			& {\color{black}KLD ($T$=1)} & {\color{black}W4/A8} & {\color{black}25} & {\color{black} 278} &  {\color{black}78.11} \\
			& {\color{black}KLD ($T$:dynamic\cite{jafari2021annealing})} & {\color{black}W4/A8} & {\color{black}25} & {\color{black} 278} &  {\color{black}78.55} \\
			& MAE & W4/A8 & 25 & {\color{black} 278} & \textbf{78.86} \\
			
			\bottomrule
		\end{tabular}
	\end{center}
\end{table}

\section{Conclusions}

In this paper, we propose an enhanced version on top of PSAQ-ViT, called PSAQ-ViT V2, which is a more accurate and general data-free quantization framework for ViTs. Specifically, we follow the relative value metric proposed by PSAQ-ViT and maximize the entropy of patch similarity to facilitate the approximation of Gaussian noise to the real images, which is also the core component of sample generation in PSAQ-ViT V2. 
On this basis, we introduce an adaptive teacher-student strategy, which facilitates the constant cyclic evolution of the generated samples and the quantized model under the supervision of the FP model. Its competitive and interactive nature can improve the informativeness
and diversity of the generated samples and encourage more effective learning of the quantized model, thus significantly improving the quantization accuracy.
It is also worth noting that the model prior and objective functions we employ are task- and model-independent, making PSAQ-ViT V2 general-purpose and compatible with various vision tasks, including image classification, object detection, and semantic segmentation.
We perform extensive experiments on various models for multiple tasks and consistently obtain SOTA results, which can even outperform the real-data-driven method under the same settings.
In addition, for a more intuitive representation, we also visualize the generated samples on the image classification task and the results of the quantized model on the object detection and semantic segmentation tasks.

As part of future work, one could combine more advanced quantization methods (\eg, non-uniform quantization) to achieve lower bit-precision (\eg, 4-bit) compression. Furthermore, one could extend the sample generation method to a wider range of application scenarios, including data-free knowledge distillation and {\color{black}data-free} black-box attacks.

\bibliographystyle{ieeetr}
\bibliography{egbib}

\end{document}